Adaptive Local Window for Level Set Segmentation of CT and MRI Liver Lesions

Assaf Hoogi[a], Christopher F. Beaulieu[b], Guilherme M. Cunha[c], Elhamy Heba[c], Claude B. Sirlin[c], Sandy Napel[d], and Daniel L. Rubin[a]

[a]Departments of Radiology and Medicine (Biomedical Informatics Research), Stanford University, Stanford, CA, USA (ahoogi@stanford.edu, dlrubin@stanford.edu)

[b]Department of Radiology and, by courtesy, Orthopedic Surgery, Stanford University, Stanford, CA, USA (beaulieu@stanford.edu)

[c]Department of Radiology, University of California, San Diego Medical Center, San Diego, CA, USA (mouracunha@hotmail.com, erheba@ucsd.edu, csirlin@ucsd.edu)

[d]Department of Radiology and, by courtesy, Electrical Engineering and Medicine, Stanford University, Stanford, CA, USA (snapel@stanford.edu)

**Corresponding Author: Assaf Hoogi**

**Abstract**

We propose a novel method, the adaptive local window, for improving level set segmentation technique. The window is estimated separately for each contour point, over iterations of the segmentation process, and for each individual object. Our method considers the object scale, the spatial texture, and changes of the energy functional over iterations. Global and local statistics are considered by calculating several gray level co-occurrence matrices. We demonstrate the capabilities of the method in the domain of medical imaging for segmenting 233 images with liver lesions. To illustrate the strength of our method, those images were obtained by either Computed Tomography or Magnetic Resonance Imaging. Moreover, we analyzed images using three different energy models. We compare our method to a global level set segmentation and to local framework that uses predefined fixed-size square windows. The results indicate that our proposed method outperforms the other methods in terms of agreement with the manual marking and dependence on contour initialization or the energy model used. In case of complex lesions, such as low contrast lesions, heterogeneous lesions, or lesions with a noisy background, our method shows significantly better segmentation with an improvement of $0.25 \pm 0.13$ in Dice similarity coefficient, compared with state of the art fixed-size local windows (Wilcoxon, p < 0.001).

## 1. Introduction

Deformable models are popular methods widely used in curve evolution applications, specifically for medical image segmentation (Caselles et al.,1997; Chan and Vese, 2001; Osher and Sethian, 1988; Li et al., 2005; Li et al., 2007; Li et al., 2008; Tsai et al., 2001; Vese and Chan, 2002). These models identify the boundary of an object, and are able to handle the challenging characteristics common in medical images, namely shape variations, image noise, intensity heterogeneities, and discontinuous object boundaries (Li et al., 2011). Level sets are a non-parametric deformable model, and thus can handle topological changes during curve evolution (Krishnamurthy et al., 2004; Li et al., 2011; Smeets et al., 2008; Tan et al., 2013; Yim et al., 2003).

Level set methods include both edge-based (Caselles et al., 1997; Li et al., 2005; Malladi et al., 1995; Osher and Sethian, 1988) and region-based models (Chan and Vese, 2001; Lankton et al., 2007; Lankton and Tannenbaum, 2008; Li et al., 2007; Li et al., 2008; Mumford and Shah, 1989; Ronfard, 2002; Tsai et al., 2001; Vese and Chan, 2002;). Edge-based models are not ideal for noisy images, objects with incomplete boundaries, or objects with low object-to-background contrast. Region-based models estimate spatial statistics of image regions to find the minimal energy where the model best fits the image. Chan and Vese (2001) applied a region-based segmentation model with global constraint, based on the Mumford and Shah functional. The main advantage of the global constraint is its high robustness to the location of the initial contour (Chan and Vese, 2001; Yezzi et al., 2002). However, in many cases, such as with heterogeneous intensity areas, a local framework performs better than a global one (Lankton and Tannenbaum, 2008; Li et al., 2007; Li et al., 2008; Malladi et al., 1995; Zheng et al., 2013; Zhang et al., 2010; Wang et al., 2009). Hybrid models are superior, defining an energy functional with local and global constraints to obtain a more accurate segmentation that is more robust to contour initialization (Smeets, et al., 2008; Zhang et al., 2010). Those methods allow curve deformation to find only significant local minima and delineate object borders despite noise, poor edge information, and heterogeneous intensity profiles (Lankton et al., 2007).

Robustness to initial conditions is an important measure of segmentation performance. Initial shape and position parameters (size, rotation, and location) need to be adequately determined; otherwise, the contour may converge to a local minimum and fail to capture the features of interest. The most common techniques for contour initialization are 1) manual selection of initial points (Ardon and Cohen, 2006; Cohen and Kimmel, 1997; Neuenschwander et al., 1994); 2) analysis of the external force field (Ge and Tian, 2002; He et al., 2006; Liu and Fox, 2005; Tauber et al., 2005); 3) naive geometric models such as a circle in 2-D or sphere in 3-D; and, 4) learned shape priors, where a statistical shape model is estimated, and the automated procedure then tries to find the segmentation that best fits the shape model (Cootes et al., 1999; Das et al., 2006; Freedman and Zhang, 2005; Tsai et al., 2003). However, methods based on shape priors may be restrictive in applications involving highly variable shapes.

Though the accuracy of contour initialization is important, the size of the local window surrounding each contour point plays a key role in the segmentation performance. The window size defines how the local scale of the statistics evaluation, and thus must be selected appropriately, even when initialization of the active contour is relatively accurate. Furthermore, well-defined local window can compensate on low-quality initial contour. Most local segmentation methods use candidates for pre-defined window sizes as input. Each candidate window size is tested over an extensive sequence of images to ascertain the best window size and this constant window size is used for the entire database of images. However, this window size will not be optimal for all images and new images with different spatial statistics may require additional experiments to find the best window size. Thus, choosing a fixed window size by trial and error is a time consuming process. Moreover, when the images contain substantial diversity of spatial characteristics, pre-defining a single window size may result in non-optimal segmentation performance for all images. For that reason, a varied window size that is defined adaptively according to spatial information has a greater chance of providing accurate segmentation.

An et al. (2007) implemented a local framework at two different scales and show that segmentation performance is better when using more than one scale. Li et al. (2008) studied in-depth the selection of kernel functions and its effect on segmentation performance. The authors applied their method using three different predefined Gaussian scales. The most accurate segmentation was obtained by using the smallest scale, but their method was more robust to contour initialization using a larger scale (Li et al., 2008), creating a trade-off when choosing a local scale. For those multi-scale methods (An et al., 2007; Li et al., 2008), a pre-specified pyramid of discrete Gaussian scales should be supplied as input. Using pre-specified scales may lead to a high dependence of segmentation accuracy on the number and the values of the discrete scales input.

Recently published research provides methods to select the best scale from a range of input scales. Yang and Boukerroui (2011) proposed a Gaussian scale selection based on the intersection of confidence intervals rule: the local scale is estimated by minimizing the mean square error of a local polynomials approximation (LPA). Pivano and Papadopoulo (2008) also applied a given pyramid of Gaussian kernels to compute local means and variances. They recommend choosing a scale that is the smallest one that induces an evolution speed greater than a user-determined threshold (Piovano and Papadopoulo, 2008). Choosing a single scale has some drawbacks. First, it may be sensitive to the criteria used. Second, since scale choice is done by examining a specific scale, it is based on the local window only; however, in many cases such as low contrast objects, heterogeneous objects or objects with a noisy background, global information can contribute substantially to segmentation performance and thus should be considered when a scale is chosen. Finally, these methods use the same chosen single scale during the whole segmentation process.

We propose a method to overcome these limitations by estimating the appropriate local window size for each contour point. The local window size is re-estimated at each point by an iterative process that considers the object scale, local and global texture statistics, and minimization of the cost function, thus generating an adaptive local window. Further, this proposed method estimates the size of the local window directly from the image, not by testing a specific scale from a range of scales and thus requires no pyramid of pre-defined scales as input, removing any potential sensitivity to user input regarding scale sizes. To the best of our knowledge, this kind of method has never been described before.

We study the effect of the adaptive window size on segmentation results in-depth, and demonstrate the capabilities and strength of our method through analysis of clinically-diagnosed lesions imaged by two different imaging modalities (CT and MRI). We use three different local energy models for image analysis of these lesions to confirm the method works with a variety of models, and compare our method to results from analyses using a global segmentation and a pre-defined fixed-size square window in local segmentation framework.

In section 2, we present the global and the local energy models that are the basis for our image analysis. Section 3 presents our method for estimating the adaptive local window. In sections 4 and 5, we discuss the key ideas regarding the implementation and the experimental data. In section 6, we present the results and compare the adaptive local window model with the fixed local window model. Section 7 discusses results and provides some concluding remarks.

## 2. Energy Models

We used three different energy models to extensively evaluate our proposed adaptive local window. The piecewise constant model provides both a global and local energy frameworks. We also use the local mean separation and histogram separation energy models.

### 2.1. Piecewise constant (PC) model

Chan and Vese (2001) present a global framework of the PC model, which assumes that an image $I$ is formed by two distinct areas (object and background areas), each of which have homogeneous intensities. Let $M_u$ and $M_v$ represent the mean intensity of the object and its background, respectively. Set $\Omega$ as a bounded subset in $R^2$ and $I(x, y)$ as the coordinates of a point on image $I$. Let $\phi(x, y)$ be a signed distance map and $\nabla$ be the first variation of the energy with respect to the distance map $\phi(x, y)$. Let $F_{PC}(M_u, M_v, \phi)$ be a function that models the object and its background:

$$F_{PC}(M_u, M_v, \phi) = \mu \int_\Omega \delta\phi(x, y) |\nabla \phi(x, y)| dx\, dy +$$
$$\lambda_1 \int_\Omega |I(x, y) - M_u|^2 H\phi(x, y) dx\, dy +$$
$$\lambda_2 \int_\Omega |I(x, y) - M_v|^2 (1 - H\phi(x, y)) dx\, dy, \tag{1}$$

where $\mu$ affects the smoothness of the curve, and $H\phi(x, y)$ is given by the Heaviside function:

$$H\phi(x,y) = \begin{cases} 1, & \phi(x,y) > \varepsilon \\ 0, & \phi(x,y) < -\varepsilon \\ \dfrac{1}{2}\left\{1 + \dfrac{\phi(x,y)}{\varepsilon} + \dfrac{1}{\pi}\sin\left(\dfrac{\pi\phi(x,y)}{\varepsilon}\right)\right\}, & |\phi(x,y)| < \varepsilon \end{cases},$$

(2)

which defines the interior of C, where C is a closed contour in $\Omega$ represented by the zero level set (ZLS), $C = \{(x,y)\,|\,\phi(x,y) = 0\}$; the exterior of $C$ is defined by (1-$H\phi(x,y)$). We can represent the narrow band around the ZLS contour $C$ by taking the derivative of $H\phi(x,y)$ to obtain a smooth approximate version of the Dirac delta $\delta\phi(x, y)$. A local version of the PC model can by used by replacing $M_u$ and $M_v$ with their local versions, $m_u$ and $m_v$, to represent the local means of a region surrounding each contour point (Lankton and Tannenbaum, 2008).

## 2.2. Mean separation (MS) model

The mean separation model was first proposed by Yezzi et al. (2002). It assumes that the object and its background have maximal separation between mean intensities:

$$F_{MS}(m_u, m_v, \phi) = \mu \int_{\Omega_n} \delta\phi(x,y)\left|\nabla\phi(x,y)\right| dx\,dy +$$

$$\lambda_1 \int_{\Omega_n} \left|\frac{I(x,y) - m_u}{A_u}\right|^2 H\phi(x,y)\,dx\,dy +$$

$$\lambda_2 \int_{\Omega_n} \left|\frac{I(x,y) - m_v}{A_v}\right|^2 \left(1 - H\phi(x,y)\right)dx\,dy.$$

(3)

Here, $\Omega_n \in R^2$ is the local version of $\Omega$ that represents the narrow-band points only. $A_u$ and $A_v$ are the areas of the local interior and exterior regions surrounding a contour point, respectively:

$$A_u = \int_{\Omega_n} H\phi(x,y)\,dx\,dy, \quad A_v = \int_{\Omega_n} (1 - H\phi(x,y))\,dx\,dy.$$

(4)

The MS energy is minimized when $|m_u - m_v|$ is maximized. In some cases, the MS model supplies better results than the PC model due to its use of the maximal contrast between the interior and the exterior regions without any restrictions on intensity homogeneity within each region. This method finds image edges well without being affected by the uniformity of the interior or exterior regions.

## 2.3. Histogram separation (HS) model

The HS model also allows for heterogeneous intensities. Let $p_u(b)$ and $p_v(b)$ be two intensity histograms computed from local interior and exterior regions, respectively, that surround each ZLS contour point, where $N$ is the number of histogram bins. The Bhattacharyya coefficient $B$ is used to evaluate the similarity of the two histograms (Bhattacharyya, 1943):

$$B = \int_{b=1}^{N} \sqrt{p_u(b)p_v(b)}db.$$

(5)

Using separate intensity histograms allows the interior and exterior regions to be heterogeneous, as long as their intensity profiles differ from each other. By using the Bhattacharyya metric to quantify the separation of the intensity histograms, we can segment objects that have local non-uniform intensities. Thus, no preliminary assumption regarding the gray level distribution of the object is made; however, the intensity profile of the entire object and the entire background must be distinct. We can then model the image using:

$$F_{HS}(x,y,\phi) = \int_{\Omega_u} B\left(\frac{1}{A_u} - \frac{1}{A_v}\right)dxdy +$$

$$\lambda_1 \int_{\Omega_u} \left(\frac{1}{A_u} \times \sqrt{\frac{p_u(x,y)}{p_v(x,y)}}H\phi(x,y)\right)dx\,dy -$$

$$\lambda_2 \int_{\Omega_u} \left(\frac{1}{A_v} \times \sqrt{\frac{p_v(x,y)}{p_u(x,y)}}(1-H\phi(x,y))\right)dx\,dy,$$

(6)

where $A_u$ and $A_v$ are defined as in equation (4) and $H\phi(x,\,y)$ is given by the Heaviside function (equation 2).

## 3. Proposed method

The PC, MS and HS regional models require that a local window be defined to create a local version of the described energies in which the energy cost function can be calculated. Here, we present a method to estimate adaptively the size of the local window surrounding each contour point (Fig. 1). The process is applied iteratively for each ZLS contour point, over iterations of the energy minimization function.

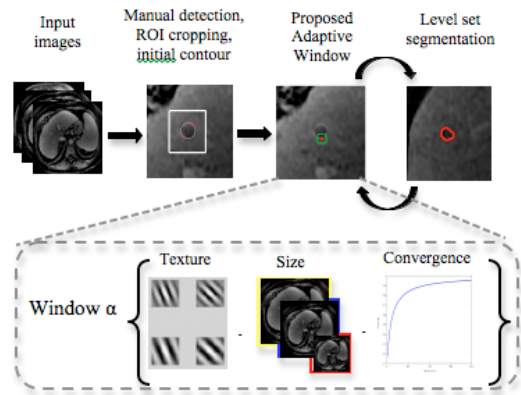

**Fig 1.** Flowchart of the main steps of the proposed method. The upper scheme presents the overall segmentation procedure. The lower scheme focuses specifically on the proposed method - calculation of the local adaptive window.

We expect that using different window sizes for different lesions and contour points will lead segmentation to fit better with changes in the spatial statistics. To best accommodate varying spatial characteristics, the window size is estimated for the X and Y dimensions separately using the texture component of the appropriate axis only. The size of the local window depends on 1) lesion size, 2) spatial texture, and 3) convergence of the energy functional over iterations (Fig. 2).

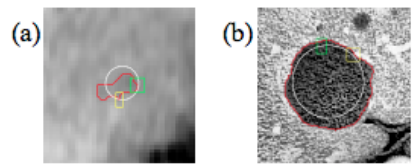

**Fig 2.** Adaptive local window sizes estimated for different lesions and different contour points. For each image, the adaptive local window size chosen is shown for two different contour points (a) Low contrast MRI liver lesion. Yellow window – 3 x 5 pixels, green window – 5 x 5 pixels. (b) Noisy CT liver lesion. Yellow window – 11 x 11 pixels, green window – 9 x13 pixels. Red contour – radiologist manual marking. White contour – initial zero level set (ZLS) contour.

### 3.1. Adaptive local window (ALW)

The adaptive local window depends on the size of the lesion to be segmented. It should be smaller for smaller lesions.

The adaptive window also depends on the spatial texture of the lesion and its background. The window should ensure convergence of the ZLS towards the lesion boundaries, even in the case of low contrast lesions, while preventing convergence of the contour into local minima in cases of a noisy background. We thus estimate the appropriate local window size using also texture analysis, done by extracting Haralick texture features, i.e. contrast and homogeneity, from a second order statistics model, the gray-level co-occurrence matrices (GLCM) (Haralick, et al., Honeycutt and Plotnick, 2008; 1973; Wang and Georganas, 2009;). For each point $(x, y)$ examined in image $I$, we compare pairs of pixels, where the second pixel in the pair is $(x + \cos\theta, y + \sin\theta)$, where $\theta \in (0, 90, 180, 270)$ degrees. Thus, each pixels' pair represents a comparison between the selected pixel and one pixel away in each of the four angular directions.

Let $W$ be a general local window of $X_W \times Y_W$ pixels, surrounding an examined contour point within a region $I$. The co-occurrence matrix $P(m, n, \theta)$ of $W$ is defined as the number of pixel pairs $(x, y)$ and $(x + \cos\theta, y + \sin\theta)$ in $W$ with grey values of $(m,n)$:

$$P(m,n,\theta) = \sum_{x=1}^{X_W} \sum_{y=1}^{Y_W} \begin{cases} 1, & I(x,y) = m \text{ and } I(x+\cos\theta, y+\sin\theta) = n \\ 0, & \text{otherwise} \end{cases}. \tag{7}$$

Then, homogeneity and contrast criteria are evaluated for each $\theta$ as:

$$\text{homogeneity}(\theta) = \sum_{m,n=0}^{N_G-1} P(m,n,\theta)(1+|m-n|^{-1})$$

$$\text{contrast}(\theta) = \sum_{m,n=0}^{N_G-1} |m-n|^2 P(m,n,\theta), \tag{8}$$

where $N_G$ is the total number of grey levels considered. These spatial criteria are averaged over all $\theta$'s separately for each axis, $X$ and $Y$. For local analysis, criteria are evaluated for each ZLS point separately. For global analysis, those criteria are calculated and averaged over all points within the lesion bounding box.

Along with the analysis of lesion size and the spatial texture, we consider the progression of energy minimization over iterations. Thus, we use lesion size, spatial texture and energy minimization to estimate the adaptive window surrounding each ZLS contour point:

$$\hat{W}_{x_{ij}} = \frac{L_x}{\log(L_x)} \left( GH + \frac{1}{GC} + \frac{1}{LC_{x_{ij}}} + \frac{1}{F_{j-1}} \right)^{-1}$$

$$\hat{W}_{y_{ij}} = \frac{L_y}{\log(L_y)} \left( GH + \frac{1}{GC} + \frac{1}{LC_{y_{ij}}} + \frac{1}{F_{j-1}} \right)^{-1}, \tag{9}$$

The local window is estimated for the $i^{\text{th}}$ ZLS contour point over the $j^{\text{th}}$ iteration. $L_x$ and $L_y$ are the X and Y lesion dimensions. To represent the effect of the lesion size appropriately, and to provide accurate segmentation for a wide range of lesion sizes, we apply a non-linear log operator, dividing $L_x$ and $L_y$ by $\log(L_x)$ and $\log(L_y)$, respectively. $GH$ is the global homogeneity and $GC$ is the global contrast; each of which are calculated once for each lesion. $LC$ is the local contrast, calculated in the X and Y directions separately. Both global and local information play an important role in determining the appropriate window size and thus both are included. $\overline{F_{j-1}}$ represents the average of the energy functional over all ZLS contour points during the previous iteration. As long as curve evolution continues, the average value of $F_{j-1}$ will decrease as the size of the local window decreases. Therefore, $\overline{F_{j-1}}$ demonstrates the progression of energy minimization over iterations.

### 3.2. Method optimization

To optimize local energies, each point is considered separately, and moves to minimize the energy computed in its own local region. Local neighborhoods are split into local interior and local exterior by the evolving ZLS curve. Energy optimization is done by using the energy model in an adapted surrounding region. Let $E(\phi)$ be an energy functional that is derived by a localization of the force $F(I, \phi)$, in our case $F_{PC}$, $F_{MS}$ or $F_{HS}$:

$$E(\phi) = \int_{\Omega_n} \delta\phi(x,y)F(I(x,y),\phi(x,y))\,dx\,dy.$$

$(10)$

The first variation of (10) is defined as:

$$E(\phi + \xi) = \int_{\Omega_n} \delta(\phi(x,y) + \xi) \times F(I(x,y),\phi(x,y) + \xi)\,dx\,dy,$$

$(11)$

where $\xi$ represents a small change along the normal direction of $\phi(x,y)$. Taking the partial derivative of (11) and considering the minor differential of the perturbation ($\xi \rightarrow 0$):

$$\nabla_{\xi|\xi=0}E(\phi + \xi) = \int_{\Omega_n} \delta\phi(x,y)\,dx\,dy \times \int_{\Omega_n} |\nabla\phi(x,y)|,\phi(x,y))\,dx\,dy + \int_{\Omega_n} \eta\phi(x,y)\,dx\,dy$$

$(12)$

where $\eta\phi(x, y)$ represents the derivative of $\delta\phi(x, y)$, which is equal to zero for every ZLS point and thus does not affect curve evolution. The Cauchy-Schwartz inequality can be used to show that the optimal direction for curve evolution is (Lankton and Tannenbaum, 2008):

$$\frac{\partial\phi(x,y)}{\partial t} = \int_{\Omega_n} \delta\phi(x,y)\nabla\phi(x,y),\phi(x,y))\,dx\,dy$$

$(13)$

We apply equation (13) to evolve the ZLS curve between sequential iterations of the segmentation process.

## 4. Implementation details

### 4.1. Image preprocessing

Normalization of gray values is done for each image separately. We apply contrast-limited adaptive histogram equalization (CLAHE) with a uniform distribution (Zuiderveld, 1994) to enhance low contrast lesions, while preventing enhancement of noise. Due to the high diversity of our dataset, both low contrast and noisy regions exist. Thus, we apply bilinear interpolation between neighboring patches to eliminate artificially induced boundaries.

### 4.2. Lesion detection and distance map reconstruction

Two board-certified abdominal imaging radiologists manually annotated all lesion boundaries by marking two points that approximate the lesion's long axis (white plus signs, Fig. 3). These points are used to create a region of interest (ROI) by taking a 10-pixels interval from those edge points. An initial zero level set (ZLS) is obtained by using those 2 points as the diameter of the circular contour (Fig. 3).

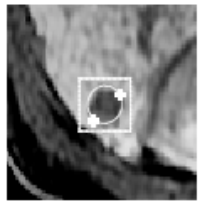

**Fig. 3.** Reconstruction of the initial contour. The radiologist marks two points to indicate a lesion's long axis (white plus signs), from which a ROI (white rectangle) is constructed. An initial circular ZLS contour is created (white circle).

### 4.3. Narrow-band

To save computation time, the proposed method calculates the energy functional only for grid points located within a narrow-band of the distance map $\phi(x, y)$ around $C$ (Fig. 4). Chopp (1993) was the first to introduce the narrow-band idea that is now commonly used in implementations of local segmentation frameworks (Lankton and Tannenbaum, 2008; Li et al., 2005; Peng et al., 1999). The segmentation process begins with initialization of every pixel within a narrow band surrounding the current contour, so that values of exterior and interior statistics are estimated (Lankton and Tannenbaum, 2008). An update of the distance map $\phi(x, y)$ then occurs within the narrow band. Thus, using the local framework, the initialization computations can be significantly reduced depending on the size of the local window and on the initial location of the contour relative to its final position (Lankton and Tannenbaum, 2008).

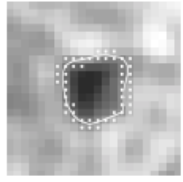

**Fig. 4.** Zero level set contour (white) with a chosen narrow band (white dots).

## 5. Experimental details

Our institutional review board approved this study. We analyzed 233 liver lesions divided into two subsets. The first subset contains 69 contrast-enhanced CT images (Siemens Medical Solutions, Erlangen, Germany) of liver lesions (20 hemangiomas, 25 cysts and 24 metastases). The following image acquisition parameters were used: 120 kVp, 140–400 mAs, 2.5-5 mm section thickness and pixel spacing of $0.704 \pm 0.085$ mm. The second subset has 164 liver lesions from MRI scans scanned at a different academic institution. All patients underwent 3T gadoxetic acid enhanced MRI (Signa Excite HDxt; GE Healthcare, Milwaukee, WI) at a tertiary liver center for evaluation of suspected hepatocellular carcinoma (HCC) and were found to have one or more LI-RADS (LR) legions classified as LR-3 or LR-4. Slice thickness was 5 mm and pixel spacing of $0.805 \pm 0.078$ mm. Pulse sequences of single-shot fast spin-echo T2-weighted, and pre- and post-contrast axial 3D T1-weighted fat-suppressed gradient-echo were used.

In the full set of 233 lesions, a wide range of lesion sizes occurred. The size of CT liver lesions ranged from $18.58 \times 20.20$ mm to $125.24 \times 132.15$ mm. The MRI liver lesions ranged from $16.87 \times 14.06$ mm to $32.81 \times 36.56$ mm. The different imaging modalities and clinical diagnoses result in high diversity of lesion characteristics present (Fig. 5). This illustrates the importance of, and need for, an adaptive local window size that is able to handle a wide range of spatial characteristics. These spatial criteria serve as key ideas for our method as given in equation (9).

Radiologists traced a single 2D slice for each liver lesion and these manual markings have been used to evaluate the proposed method. The automated segmentation contours were extracted and quantitatively compared with the average of the two radiologists' marking. To investigate sensitivity of the results to initialization, we changed the angle and length of the user-supplied linear markup. Five linear markups were generated for each lesion, by randomly adjusting the position of the two input points within a 5-pixel diameter. A 5-pixel diameter was considered reasonable, based on intra-variability of radiologists.

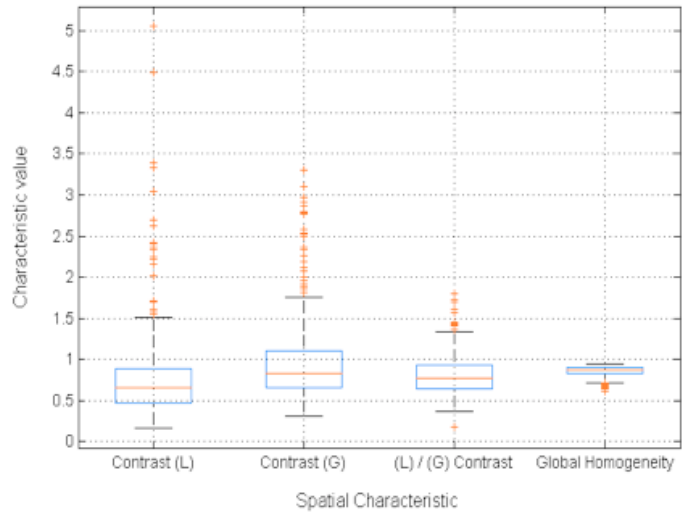

**Fig. 5.** Spatial characteristics of the lesions (L) and the whole ROI (G).

## 6. Results

### 6.1. Size distribution of the adaptive local windows

The estimated window sizes ranged from 9-35 pixels (mean ± SD: 13.65 ± 3.29 pixels) for CT liver lesions and from 5-19 pixels (8.65 ± 2.14 pixels) for MRI liver lesions. Therefore, using a single fixed-size local window for all lesions may be inaccurate and will create an inconsistent segmentation performance, means that a specific window size will be good for some lesions, but for other it will not.

### 6.2. Segmentation performance

Cost function parameters of $\mu_1 = 0.15$, $\lambda_1 = 2$, $\lambda_2 = 2$ were used, as they supplied the best average results for all 233 lesions. Figure 6 shows some examples of segmentation for different lesions. Segmentation performance was assessed using the Dice similarity coefficient (Table 1). The Dice coefficient was calculated relative to each radiologist's manual marking, and then an average Dice score estimated. Our proposed segmentation method has high agreement with the manual markings for different local energies. No significant differences were found between manual and our automated segmentations (Wilcoxon, p > 0.05).

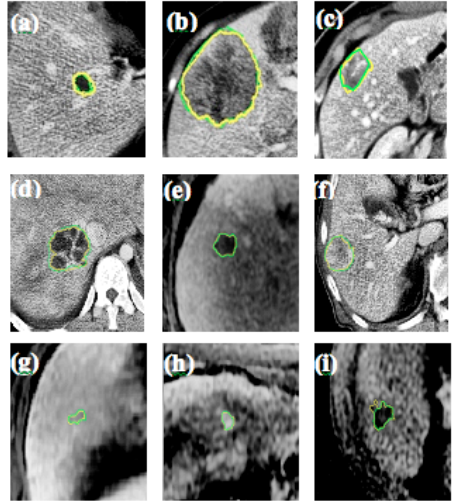

**Fig. 6.** Automatic segmentations (yellow) of liver lesions in CT (a-d, f) and MRI (e, g-i) images obtained using the ALW method with the piecewise constant model (PC). (a-b) different lesion sizes (Dice of 0.88, 0.91 respectively), (b-d) heterogeneous lesions (Dice of 0.91,0.92, and 0.89, respectively), (e) homogeneous lesion (Dice of 0.97), (f -g) low contrast lesions (Dice of 0.92, and 0.96 respectively), (h-i) noisy background (Dice of 0.93, 0.86 respectively). Green contours represent the manual annotations.

Table 1. Average dice coefficient and 95% confidence interval (CI) for the lesion analysis using the proposed adaptive local window compared to manual marking.

|  | ALW VS. Manual marking [95% CI] |
| --- | --- |
| PC energy | 0.89 [0.88, 0.90] |
| MS energy | 0.87 [0.86, 0.88] |
| HS energy | 0.88 [0.87, 0.89] |

*6.3. Process convergence*

Figure 7 demonstrates the convergence of both the Dice coefficient and the energy functional over multiple iterations. For both CT and MRI images, the Dice coefficient increased rapidly over a few iterations as the automated segmentation converged on a result that matched the manual segmentations. As expected, energy decreased with increasing iterations, converging on a single value; this implies minimization of the energy functional. For both metrics, substantial convergence was obtained after fewer than 20 iterations and there were only minor fluctuations around their final values over later iterations.

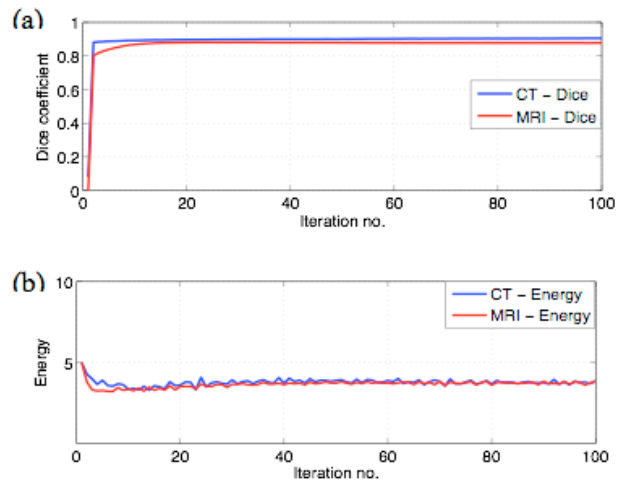

**Fig. 7.** Convergence of the (a) Dice coefficient and the (b) energy functional over iterations for both MRI and CT datasets.

### 6.4. Comparison with global energy

We compared our method with the PC global energy model of Chan and Vese (2001), using the same $\mu$, $\lambda_1$, and $\lambda_2$ parameter values. The global energy model and manual marking had a mean Dice similarity coefficient of 0.784, with a 95% Confidence Interval of [0.759-0.805]. This is significantly less than the 0.89 Dice coefficient obtained using ALW for the local PC energy model (Wilcoxon, p < 0.001). Figure 8 reveals that the global energy model shows poor agreement with manual marking when analyzing images with low contrast (Fig. 8a-8c) or heterogeneous lesions (Fig. 8d-8f), especially if the lesions are located near the boundaries of the liver itself (Fig. 8e).

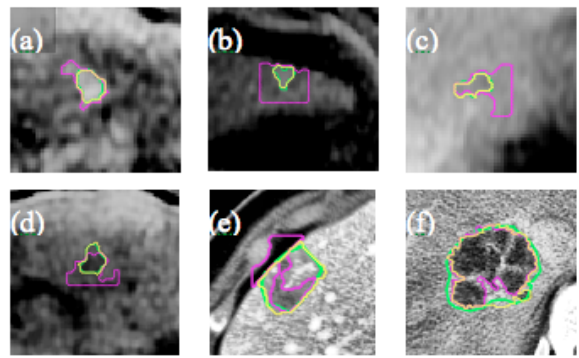

**Fig. 8.** Examples for contours that were obtained by the manual marking (green contour), the proposed ALW (yellow contour) and the global PC energy (magenta contour). Parts of the magenta contour, which are seen as a straight line, represent excessive curve evolution that was stopped by the borders of the selected ROI bounding box.

*6.5. Comparison with fixed square local window*

We also compared our ALW with state of the art methods that use a fixed square local window (FLW) surrounding each contour point. The comparison was done for each of the three local energy models – PC, MS and HS. As with the ALW and global energy models, the same parameter values ($\mu_1 = 0.15$, $\lambda_1 = 2$, $\lambda_2 = 2$) were used because they supplied the best results on average for all 233 lesions. Thus, the only difference between the FLW and the ALW methods was the size of the local window in which the statistics were calculated. Hence, any difference in performance was directly related to the local window size. For FLW, we began by testing a range of window sizes from 5x5 to 25x25 pixels. For CT lesions, 15-pixel square windows gave the best average performance, and for MRI liver lesions, 11-pixel square windows were best. Therefore, these two fixed sizes were used for ALW-FLW comparisons. For all 233 lesions, the 11-pixel FLW gave an average Dice coefficient of 0.851(95% CI: 0.84-0.86) for all three local energy models. Using a 15-pixel FLW, the mean Dice coefficient was 0.83 (95% CI: 0.81-0.85). For each applied energy model, the performance of ALW was significantly better than FLW (Wilcoxon, p < 0.01). Figure 9 shows six different lesions that demonstrate some segmentation challenges and reveal that the our ALW can handle these diverse types of images better than local FLW segmentation. The superiority of the adaptive method can be seen for each of the three energy models tested (PC, MS and HS).

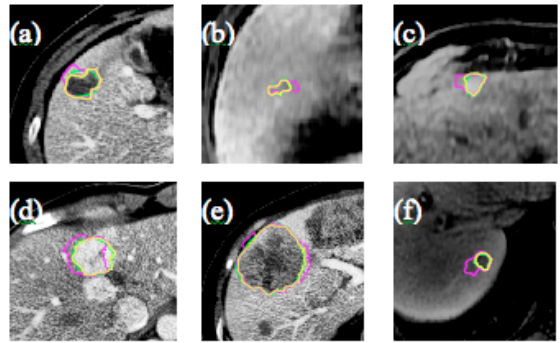

**Fig. 9.** Examples for contours that were obtained by the manual marking (green contour), the proposed ALW (yellow contour) and the FLW method (magenta contour). (a) PC model – CT liver lesion, (b) PC model – MRI liver lesion, (c) Mean separation (MS) model – MRI liver lesion, (d) Mean separation (MS) model – CT liver lesion (e) Histogram separation (HS) model – CT liver lesions, (f) Histogram separation (HS) model – MRI liver lesions.

Lesions for which one or more automated methods had less than 70% agreement with the manual marking were also examined separately (Fig. 10). Significant Dice improvement of 0.19 ± 0.09 was obtained in this subset by using our ALW (Wilcoxon, p < 0.05 for PC model, p < 0.001 for MS and HS models), compared with the classic FLW. A second subset of lesions that had performance differences between ALW and FLW of greater than 10% was considered. The 10% difference was chosen according to clinical decision, wherein ALW could be better or worse than FLW. Table 2 shows that our ALW outperforms FLW for each of the local energy models, with a significant mean performance improvement of $0.25 \pm 0.13$ (Wilcoxon, $p < 0.001$ for all three energy models). Moreover, ALW shows lower sensitivity and smaller dependence on a specific local energy than FLW.

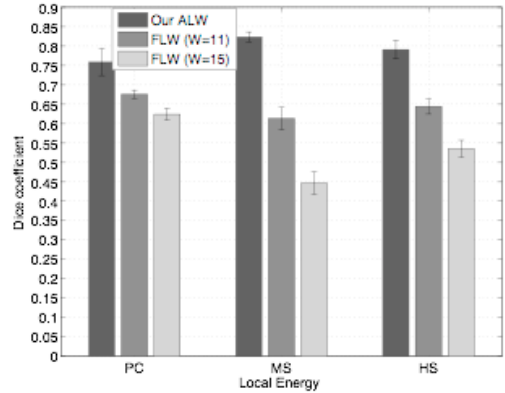

**Fig. 10.** Subset of lesions, for which one or more automated methods obtained less than 70% agreement with the manual marking. For FLW, 11-pixels and 15-pixels window sizes were used. Comparison was performed for all PC, MS and HS local energies.

Table 2. Subset of lesions for which absolute performance differences higher than 10% between ALW and FLW was obtained. Dice coefficient is presented. Wilcoxon paired test was performed between the ALW and each FLW ($p < 0.001$).

| | Number of Lesions **ALW** [95% CI], **FLW11** [95% CI] | Number of Lesions **ALW** [95% CI], **FLW15** [95% CI] |
|---|---|---|
| PC | 11 **0.84** [0.77, 0.89] **0.70** [0.63, 0.8] | 18 **0.83** [0.79, 0.87] **0.72** [0.64, 0.8] |
| MS | 42 **0.84** [0.79, 0.85] **0.54** [0.47, 0.59] | 60 **0.85** [0.81, 0.86] **0.47** [0.41, 0.52] |
| HS | 25 **0.82** [0.80, 0.85] **0.60** [0.55, 0.64] | 32 **0.83** [0.78, 0.86] **0.59** [0.54, 0.64] |

*6.6. Sensitivity to parameters*

Only one parameter is required for the decision of the local window size - the distance between each pair of pixels used for calculating the GLCM matrices. We tested distances of 1, 2, 3, 4 and 5 pixels before choosing the 1-pixel distance as best. All 233 lesions and all tested distances combined had an average Dice coefficient of $0.887 \pm 0.004$. The negligible standard deviation indicates that the segmentation performance is affected very little by the value of this parameter. Thus, the performance of our method is stable, and the parameter does not need to be re-tuned manually for different populations of lesions.

*6.7. Sensitivity to initialization*

We assessed the ability of our model to deal with deviations in the positioning of the initial contour by applying five randomly different contour initializations. Results demonstrate that the ALW method supplied the highest Dice similarity values, compared with both FLW fixed window sizes, when compared with the manual marking. Moreover, the ALW showed the smallest changes in the segmentation performance when it was applied using different local energy models, significantly better than FLW windows (Wilcoxon, $p < 0.05$ for PC model, $p < 0.001$ for MS and HS models). Table 3 shows the average performance of ALW and FLW when using five different contour initializations

Figure 11 presents two representative images to show this result visually. Each image contains two inaccurate long-axis points that lead to reconstruction of an inaccurate initial contour. The results reveal that despite with noise (Fig. 11a) and low contrast lesions (Fig. 11b), our model can deal with substantial deviations of the location of the initial contour.

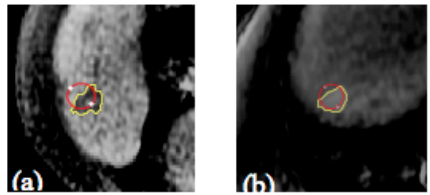

**Fig. 11.** Inaccurate initialization and the final segmentation. Both long-axis input points (white stars) and the initial contour (red circle) that was reconstructed due to those points are presented. The yellow contour in each image illustrates the final accurate segmentation of the lesion.

Table 3. Segmentation performance (mean Dice coefficient for comparison with manual marking) for the proposed ALW and the classic FLW at two different fixed sizes: 11-pixels square (FLW11) and 15-pixels square (FLW15). Results from five different contour initializations were averaged. Three different energy models were used.

|  | PC (N=233) [95% CI] | MS (N=233) [95% CI] | HS (N=233) [95% CI] |
|---|---|---|---|
| **ALW** | **0.885** | **0.86** | **0.87** |
|  | [0.88, 0.89] | [0.85, 0.86] | [0.87, 0.88] |
| FLR11 | 0.878 | 0.82 | 0.85 |
|  | [0.87, 0.88] | [0.81, 0.83] | [0.84, 0.87] |
| FLR15 | 0.88 | 0.78 | 0.83 |
|  | [0.87, 0.88] | [0.77, 0.80] | [0.80, 0.86] |

*6.8. Computational time*

We examined the computational time required by the proposed method to analyze each lesion. All data were processed using MATLAB R2013b 64–bit (MathWorks Inc., 2013). Compared to the time required by the FLW method, or proposed ALW model took 2-5 times (~20-60 s) longer to run on the same hardware.

## 7.  Discussion and conclusions

We present a novel method for adaptive re-estimation of the local window (ALW) for level set segmentations. The window is re-estimated 1) for each contour point, 2) separately for X and Y window dimensions, 3) over level set segmentation iterations, and 4) separately for each lesion in the dataset. The size of the local window is re-estimated based on 1) the size of the lesion, 2) the spatial texture, and 3) the minimization of the energy functional over iterations. The rationale for choosing those criteria is the substantial diversity of these lesion characteristics (Fig. 5).

Our method contains a local contrast term in addition to global criteria for contrast and homogeneity. The inclusion of both local and global criteria has an important benefit. We demonstrate that incorporating both global and local criteria correctly identifies the best local window size.

Our proposed method shows high agreement with expert manual marking for a diverse dataset of CT and MRI images (Fig. 6). The variety of spatial texture characteristics in our datasets emphasizes the strength of our adaptive method. Our ALW method performed well with low contrast images, heterogeneous lesions, and with noisy lesions or noisy backgrounds.

We compared our results to a global PC energy model (Fig. 8) and to models using a pre-defined fixed local window size (Fig. 9). Lankton and colleagues (Lankton et al., 2007; Lankton and Tannenbaum, 2008) note that their FLW model has high chances of breaking down in a set of widely varying lesions. Our results confirm that it is almost impossible to choose single fixed-size local window that will best fit every case (section 6.5). Among all fixed window sizes tested, an 11-pixel square window was better for MRI lesions and a 15-pixel square window was better for CT lesions. Estimating an adaptive local window using our method, results in a better agreement with traditional manual marking.

We analyzed two subsets of lesions based on results. Table 2 shows that for lesions with Dice differences higher than 10% between ALW and FLW, our ALW was significantly better. Dice similarity to manual marking was greater by $0.25 \pm 0.13$ (Wilcoxon, $p < 0.001$ for all 3 energy models). The second subset of lesions were those in which one or more automated methods obtained less than 70% agreement with the manual marking (Fig. 10). For this subset, the Dice coefficient with manual marking was improved by $0.19 \pm 0.09$ using ALW versus FLW (Wilcoxon, $p < 0.05$ for PC model, $p < 0.001$ for MS and HS models).

The method was also evaluated using five different contour initializations. Our ALW method performs better under these conditions than the FLW methods as it was affected less by different initial contour values (Table 3). The use of both local and global statistics in the model increases the segmentation agreement with the manual marking and by lowers dependence of the model on changes in the location of the initial contour.

We also examined model sensitivity to parameters. Sensitivity to the single pre-defined parameter in the method - the distance between each pair of pixels used for calculating the GLCM matrices - was very low. Therefore, parameter tuning is not needed for the evaluation of the adaptive local window.

The presented work has some limitations. First, although our database of images of 233 lesions (69 CT, 164 MR) comes from two academic institutions, a larger sample size is desirable. Second, additional manual markings for each lesion may result in a more accurate evaluation of the automated segmentation.

Future work may include an automated evaluation of the level set parameters $\mu$, $\lambda_1$, and $\lambda_2$. Extension of the method to 3D is also a future direction, as well as incorporation of automatic detection of the lesions prior to segmentation, so that the entire segmentation process is fully automated, with no dependence on user input.

In summary, the method presented shows significantly more accurate lesion segmentation than current state of the art level set methods. It performed better than prior methods in all tested configurations, including different local energy models, different contour initializations, and for different levels of lesion complexity.

**Acknowledgements**

This project was supported by the National Cancer Institute, National Institutes of Health, under Grants U01CA142555 and R01CA160251.